\documentclass[]{gewu}

\usepackage{wasysym}

\usepackage[toc,page,header]{appendix}

\usepackage{minitoc}
\usepackage{microtype}      
\usepackage[dvipsnames,table]{xcolor}         

\usepackage{hyperref}
\usepackage{url}
\usepackage{booktabs}
\usepackage{multirow}
\usepackage{subcaption}
\usepackage{caption}
\usepackage{amssymb}
\usepackage{graphicx}
\usepackage{amsmath}
\usepackage{wrapfig}
\usepackage{enumitem}
\usepackage{arydshln} 
\usepackage{gensymb}
\usepackage{makecell}
\usepackage{pifont} 
\usepackage{hyperref}
\usepackage{url}
\usepackage{xcolor}
\usepackage{booktabs}
\usepackage{multirow}
\definecolor{headercolor}{gray}{0.80}
\definecolor{gaincolor}{RGB}{0,102,204}
\usepackage{graphicx}
\usepackage{subcaption}
\usepackage{tabularx}
\usepackage{wrapfig}

\title{Gestalt: Large Multimodal Interplay Model}

\makeatletter
\renewcommand\author[2][]{%
  \addtolist[#1]{#2}{\authorlist}{\authorformat}{}%
}
\makeatother

\makeatletter

\renewcommand\author[2][]{%
  \addtolist[#1]{#2}{\authorlist}{\authorformat}{, }%
}

\newcommand\authornewline[2][]{%
  \addtolist[#1]{#2}{\authorlist}{\authorformat}{\protect\\}%
}

\makeatother

\author[1,2,\dagger,*]{Zequn Yang}
\author[1,2,\dagger,*]{Yu Miao}
\author[3,\dagger,*]{Haotian Ni}
\author[1,2,\dagger]{Ziheng Chen}

\authornewline[4\dagger]{Chengxiang Huang}
\author[5]{Dongzhan Zhou}
\author[5]{Kai Chen}
\author[5]{Qi Zhang}

\authornewline[1,2]{Ji-Rong Wen}
\author[1,2,\ddag,*]{Yake Wei}
\author[\textnormal{1,2,6,\ddag, \Letter}]{Di Hu}

\affiliation[1]{Gaoling School of Artificial Intelligence, Renmin University of China}
\affiliation[2]{Beijing Key Laboratory of Research on Large Models and Intelligent Governance}
\affiliation[3]{Beihang University}
\affiliation[4]{Beijing University of Posts and Telecommunications}
\affiliation[5]{Shanghai Artificial Intelligence Laboratory} 
\affiliation[6]{AresoX}

\makeatletter
\renewcommand\contribution[2][]{%
  \addtolist[#1]{#2}{\contributionlist}{\contributionformat}{}%
}
\makeatother

\contribution[]{$^\dagger$Equal contribution,}
\contribution[]{$^\ddag$Team leader,}
\contribution[\textnormal{\Letter}]{Corresponding author\protect\\[-2pt]}
\contribution[*]{Work partially done at Shanghai Artificial Intelligence Laboratory, as an internship.}

\abstract{
In this paper, we propose \textbf{Gestalt}, a new paradigm of large multimodal model built around multimodal interplay. Despite rapid advances, large multimodal models are reaching a bottleneck: existing approaches focus primarily on accommodating additional modalities while overlooking the distinct characteristics of each modality and the relations among them. Motivated by the multistage property of human multisensory perception, we propose a multimodal interplay pyramid that organizes multimodal modeling as a progression from modality-specific processing, through cross-modal alignment, to deeper multimodal integration. Guided by this pyramid, Gestalt adopts a unified discrete diffusion framework and an interplay-partitioned architecture, with learnable interplay tokens mediating cross-modal exchange and integration. The pyramid also structures its data organization and training strategy. Strong performance across image generation, multimodal understanding, and text-only evaluation shows that Gestalt significantly improves cross-modal integration while preserving modality-specific information, effectively harnessing the strengths of diffusion-based multimodal models and offering a promising path toward unified multimodal intelligence. 
}

\MyEmail{Zequn Yang at \email{zqyang@ruc.edu.cn}, Di Hu at \email{dihu@ruc.edu.cn}}
\checkdata[Project Page]{\url{https://GeWu-Lab.github.io/Gestalt}}
\checkdata[Code]{\url{https://github.com/GeWu-Lab/Gestalt}}
\checkdata[Model]{\url{https://huggingface.co/GeWu-Lab/Gestalt}}

\begin{document}
\maketitle

\section{Introduction}
Modern large multimodal models have evolved from architectural compatibility to data compatibility. Early systems, such as LLaVA~\citep{liu2023visual} and Qwen2.5-VL~\citep{bai2025qwen25vltechnicalreport}, achieve architectural compatibility by connecting modality-specific encoders to a pretrained language model through learned projectors. More recent native models, such as Kimi K3~\citep{kimiteam2026kimik3openfrontier} and Qwen 3.8, pursue data compatibility by jointly pretraining on large-scale mixed-modal corpora within a shared model. Yet progress along this path is reaching a bottleneck. We argue that multimodal intelligence requires more than accommodating multiple modalities: \emph{more is different}, and merely bringing modalities together cannot adequately address the emergent properties and distinctive challenges inherent to multimodality.

\begin{figure}[t]
\vspace{-2em}
  \centering
  \includegraphics[width=\linewidth]{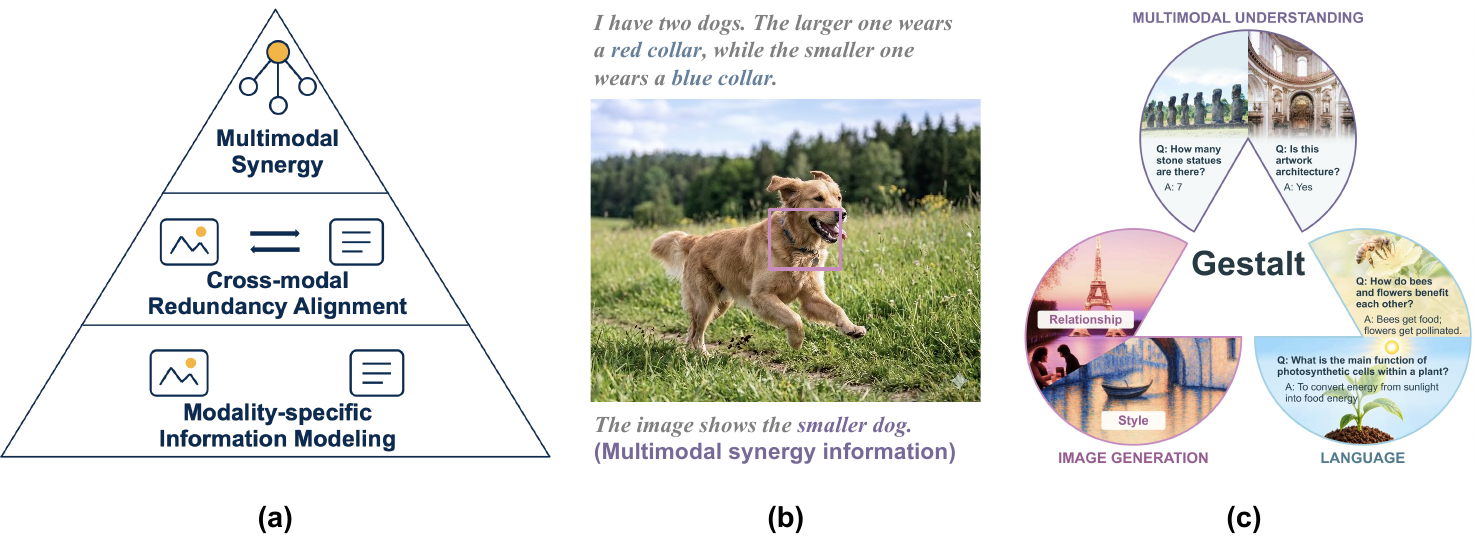}
\vspace{-2em}
\caption{\textbf{(a)} The multimodal interplay pyramid progresses from modality-specific modeling to cross-modal alignment and synergy. \textbf{(b)} Combining the textual size-collar relation with the visible collar identifies the depicted dog as the smaller one. \textbf{(c)} Gestalt supports multimodal understanding, image generation, and language tasks within a unified model.}
\vspace{0em}
  \label{fig:interplay}
\end{figure}

In multimodal settings, information may be available in one modality, shared across modalities, or emerge only from their combination. Such information relationships define multimodal interplay, a feature unique to multimodal learning. Human multisensory perception offers a useful analogy. Sensory information is first encoded within individual pathways and then progressively integrated across multiple stages into a coherent representation~\citep{stein1993merging, gazzaniga2011cognitive, tang2016interactions}. Inspired by this multistage organization, we propose a multimodal interplay pyramid that arranges multimodal learning into successive levels (\autoref{fig:interplay}a), with each level building on the preceding one to enable increasingly complex interplay.

The pyramid handles these information relationships in a bottom-up progression. The \emph{lower stage} preserves unique information available from only one modality. Building on this foundation, the \emph{intermediate stage} aligns redundant information shared across modalities. At the top, the \emph{upper stage} integrates information across modalities to derive synergistic information unavailable from either modality alone. \autoref{fig:interplay}b illustrates this progression within a single image-text pair. The text contributes the size--collar relation, whereas the image supplies the visible collar color. Their shared description of a Golden Retriever on grass establishes cross-modal correspondence, and combining the two modality-specific cues identifies the depicted dog as the smaller one.


Based on this principle, we propose \textbf{Gestalt}\footnote{The name is inspired by Gestalt psychology~\citep{kohler1967gestalt}: the whole is greater than the sum of its parts.}, a large multimodal model that explicitly accounts for different forms of information interplay across modalities. Gestalt implements this idea through its architecture, data organization, and training strategy. Architecturally, Gestalt adopts a discrete diffusion framework that represents images and text in a unified token space and trains them with the same objective, enabling both multimodal understanding and generation. To implement the interplay pyramid, the Transformer is divided into two successive zones, with learnable interplay tokens mediating cross-modal information flow throughout. In \emph{Interplay Zone I}, image and text tokens are processed separately by modality-specific feed-forward networks, and direct attention between them is restricted; cross-modal exchange is instead mediated by the interplay tokens. In \emph{Interplay Zone II}, this restriction is removed, allowing image, text, and interplay tokens to interact through full multimodal self-attention and shared feed-forward processing. The two zones thus instantiate the pyramid as a progression from modality-specific processing, through controlled exchange, to full multimodal integration.


Beyond architecture, the interplay pyramid also guides data organization and the training strategy. The modality-unique foundation of its lower stage is already established by the pretrained backbone. Multimodal training therefore begins with cross-modal alignment: Gestalt conducts joint multimodal pretraining and conditional continual pretraining on 78M aligned image-caption pairs to learn a joint vision-language distribution and bidirectional cross-modal dependencies. It then performs interplay-oriented supervised fine-tuning on 13.7M instruction examples organized into redundant, text-unique, visual-unique, and synergistic categories. Across three curriculum phases, all four categories are retained while the sampling emphasis progressively shifts from consolidating cross-modal alignment and modality-specific capability, and finally to deeper multimodal synergy.

A unified multimodal model is expected to do more than support generation and understanding. It should also build coherent cross-modal representations. This calls for a broad set of evaluations. On benchmarks that emphasize complex, fine-grained generation, Gestalt outperforms both generation-only and unified baselines, reaching $73.30$ on UniGenBench and $74.57$/$78.86$ on the short/long settings of TIIF-Bench. Gestalt also demonstrates strong multimodal understanding, particularly on vision-centric benchmarks, scoring $78.9$ and $86.1$ on the two CV-Bench tracks and outperforming the representative autoregressive-based model BAGEL on both. Representation analyses further show that Gestalt forms an integrated vision-language space while largely preserving pretrained linguistic geometry. Despite multimodal adaptation, it maintains strong language capability and delivers leading overall text-only performance among the evaluated diffusion-based unified models. Finally, results on visual-specific and synergy-oriented benchmarks, together with the organization of interplay-token representations, suggest that Gestalt adapts information exchange to different interplay demands. As illustrated in \autoref{fig:interplay}c, Gestalt establishes leading performance among the evaluated diffusion-based unified models, and it effectively harnesses the potential of diffusion-based multimodal models, offering a promising path toward unified multimodal intelligence.

\begin{figure}[t]
  \centering
  \includegraphics[width=1.0\linewidth]{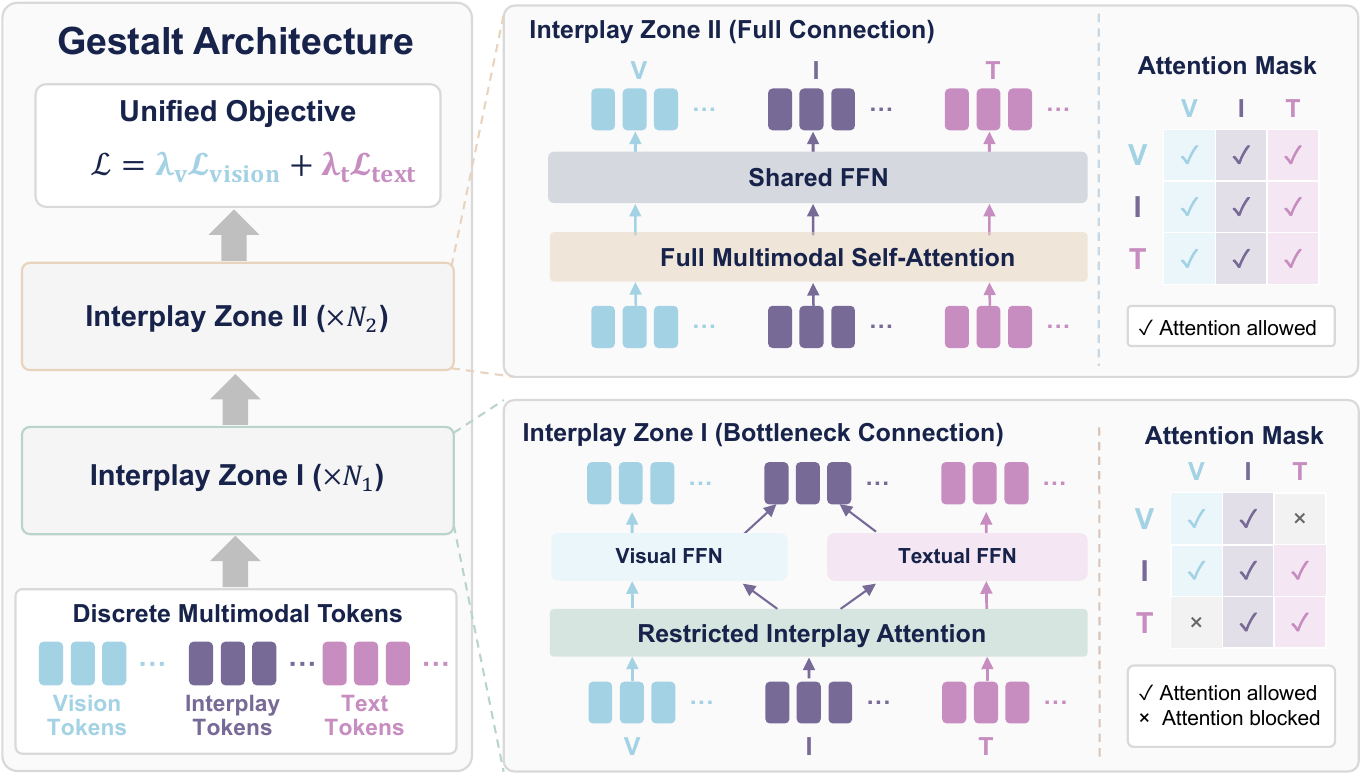}
  \vspace{0em}
  \caption{\textbf{Overview of the Gestalt architecture.} Gestalt processes discrete vision (orange), interplay (green), and text (blue) tokens through Interplay Zone I ($N_1$ bottleneck interplay layers) and Interplay Zone II ($N_2$ full multimodal interplay layers) under a unified objective. Zone I uses restricted interplay attention, with cross-modal exchange mediated by interplay tokens, together with modality-specific FFNs; Zone II uses full multimodal self-attention and a shared FFN.}
  \label{fig:architecture}
\end{figure}

\section{Methodology}
Guided by the interplay pyramid, Gestalt organizes multimodal learning as a progression from modality-specific processing to cross-modal alignment and multimodal synergy. Within a unified discrete diffusion framework, this principle shapes its interplay-centric architecture, interplay-oriented data organization, and progressive training strategy.
\subsection{Unified Interplay-Centric Architecture}

As illustrated in \autoref{fig:architecture}, Gestalt represents vision and text within a shared discrete token space, enabling multimodal understanding and generation in a unified framework. To explicitly organize cross-modal information exchange, we introduce learnable interplay tokens and structure the architecture into two hierarchical zones, progressively transitioning from restricted bottleneck interplay to full multimodal interplay. Together, these designs establish a preserve-then-integrate pathway for unified multimodal modeling.


\noindent\textbf{Discrete Multimodal Representation.}
To support multimodal interplay within a single model, Gestalt maps visual and textual inputs to a unified codebook and embedding space. This fully discrete formulation allows both modalities to be optimized under the same objective. The underlying masked diffusion language model uses bidirectional attention, allowing visual and textual tokens to condition on each other without the fixed direction imposed by autoregressive models. A fixed VQ-VAE encoder ~\citep{shi2025scalable} converts each image into discrete tokens using a codebook of 16,384 entries. These visual tokens are then combined with textual tokens in a shared sequence.
  
\noindent\textbf{Interplay Tokens.}
A shared discrete space enables joint modeling but does not explicitly organize cross-modal information exchange. We therefore introduce dedicated learnable interplay tokens as a shared interface between the visual and textual streams, whose function is progressively expanded from mediated exchange in Zone I to full multimodal integration in Zone II.

\noindent\textbf{Zone I: Bottleneck Interplay.}
Unique information depends on the distinct characteristics of each modality, yet these characteristics can be diluted by fully shared processing. Zone I therefore adopts a bottleneck interplay design that combines modality-specific modeling with controlled cross-modal exchange. Visual and textual tokens pass through separate feed-forward networks, while restricted attention prevents direct interaction between them. Instead, cross-modal information is exchanged through the interplay tokens. Through this mediated exchange, Zone I preserves modality-specific representations while gradually establishing cross-modal correspondence.

\noindent\textbf{Zone II: Full Multimodal Interplay.}
After modality-specific representations and basic cross-modal correspondence have been established, Zone II replaces the bottleneck with full multimodal self-attention and a shared feed-forward network. Visual, textual, and interplay tokens can now attend to one another directly, enabling information exchange across the entire sequence. The interplay tokens continue to aggregate task-relevant information from both modalities. Together, full multimodal attention and shared processing support the modeling of synergistic information.

\subsection{Training Data Organization}
\label{sec:training_recipe}

Gestalt organizes training data progressively across pretraining, continual pretraining (CPT), and supervised fine-tuning (SFT). For both pretraining and CPT, we use 78M aligned image--caption pairs to learn basic cross-modal correspondence and acquire multimodal knowledge.  For SFT, we organize instruction data according to the interplay pyramid. Each example is categorized as redundant, text-unique, visual-unique, or synergistic. The resulting corpus contains 13.7M examples, including 2.8M redundant, 4.2M text-unique, 4.0M visual-unique, and 2.7M synergistic examples. This organization enables controlled learning of different forms of multimodal interplay during fine-tuning. A detailed description of our data organization strategy is provided in \autoref{sec:data_organization}.

\vspace{-0.5em}
\subsection{Progressive Training Strategy}

Training follows the interplay pyramid through joint multimodal pretraining, conditional continual pretraining, and interplay-oriented supervised fine-tuning under a unified masked-diffusion framework. Because Gestalt is initialized from a pretrained backbone, its modality-specific foundation is already established before multimodal training. The multimodal stages therefore begin by aligning cross-modal redundant information, then strengthen the modeling of visual-unique information, and finally advance to cross-modal synergy.

\noindent\textbf{Unified Masked Diffusion Objective.}
With vision and text represented as discrete tokens, Gestalt jointly models both modalities with a unified masked-diffusion objective. For a multimodal sequence $\mathbf{x}=[\mathbf{x}^{v},\mathbf{x}^{t}]$, we randomly mask a set of positions and reconstruct the corresponding tokens:
\begin{equation}
  \mathcal{L}_{\text{mask}}
  = -\mathbb{E}_{m}\sum_{i\in m}
  \log p_{\theta}(x_i \mid \mathbf{x}_{\setminus m}),
\end{equation}
where $m$ denotes the set of masked positions. Unlike autoregressive modeling, this enables bidirectional conditioning across modalities without a fixed generation order.

\noindent\textbf{Progressive Multimodal Pretraining.}
Multimodal pretraining begins with joint masked prediction. Visual and textual tokens are masked independently and reconstructed jointly:
\begin{equation}
p(\mathbf{x}^{v}_{m_v},\mathbf{x}^{t}_{m_t}
\mid
\mathbf{x}^{v}_{\setminus m_v},
\mathbf{x}^{t}_{\setminus m_t}).
\end{equation}
This objective encourages the model to capture both intra-modal structure and cross-modal dependencies in either direction. This stage establishes a basic joint distribution over the two modalities.

\begin{wrapfigure}{r}{0.35\textwidth}
    \centering
    \includegraphics[width=\linewidth]{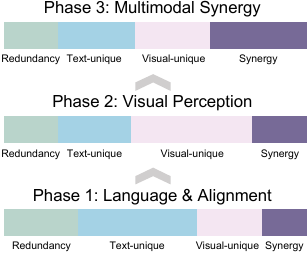}
    \caption{Training recipe for interplay-oriented data organization during SFT.}
    \label{fig:recipe}
\end{wrapfigure}

Continual pretraining then adopts conditional masked prediction: one modality remains visible while the other is masked and predicted. Learning bidirectional vision-to-text and text-to-vision dependencies strengthens cross-modal conditional modeling and provides a natural bridge between joint multimodal modeling and downstream generation conditioned on either modality.

\noindent\textbf{Interplay-Oriented Supervised Fine-tuning.}
Building on the cross-modal dependencies learned during pretraining and CPT, supervised fine-tuning focuses on teaching the model how to exploit different types of multimodal information for downstream tasks. To this end, we adopt a three-phase curriculum under conditional masked prediction. All four interplay categories are retained throughout training, while their sampling ratios are progressively adjusted to place increasing emphasis on visual-unique and synergistic information, as illustrated in Figure~\ref{fig:recipe}. This three-phase curriculum gradually shifts the learning focus from consolidating cross-modal alignment and language capability, to strengthening visual-specific capability, and finally to promoting deeper multimodal integration.

\section{Experiment}
\subsection{Training and Evaluation Overview}

\noindent\textbf{Training and Model Configuration.}
Gestalt is initialized from  LLaDA-8B-Instruct~\citep{nie2026large}. Images are encoded into discrete tokens using IBQ~\citep{shi2025scalable}, with its visual codebook of 16,384 added to the language vocabulary, while interleaved MRoPE preserves their two-dimensional spatial positions. We use 32 interplay tokens and designate the first four Transformer layers as Interplay Zone I. The newly introduced visual FFN in each of these layers is initialized from the corresponding text FFN in LLaDA. Training proceeds in three stages: joint pretraining on 70M image-caption pairs from LLaVA-OneVision-1.5~\citep{an2025llava}, conditional continual pretraining on 8M image-text pairs, and supervised fine-tuning on 13.7M interplay-organized instruction examples from LLaVA-OneVision-1.5, MAmmoTH-VL~\citep{guo2025mammoth}, FineVision~\citep{wiedmann2025finevision}, and InternVL~\citep{wang2025internvl3}. The final stage additionally includes 2.5M text-to-image examples collected from text-to-image-2M~\citep{zou2026advancing}, ShareGPT-4o-Image~\citep{chen2025sharegpt}, and BLIP3o-60k~\citep{chen2025blip3}. All the parameters are trainable at all stages.

\noindent\textbf{Evaluation Overview.} Our evaluation proceeds from what Gestalt achieves to how its multimodal representations and interactions are organized. We begin by establishing its unified capabilities across fine-grained text-to-image generation and multimodal understanding. We then look beneath these outcomes to characterize the joint vision-language space and the retention of pretrained linguistic structure and text-only capability. Finally, we focus on cross-modal interplay, examining performance under modality-unique and synergistic information settings together with the information encoded by the interplay tokens. This progression links benchmark performance with representation geometry and internal information exchange, yielding a unified empirical account of Gestalt. Further analyses in \autoref{sec:additional_results} examine interplay-oriented curriculum learning, adaptation to remote-sensing tasks, and qualitative results on fine-grained image generation.




\begin{figure}[t]
\centering
\includegraphics[width=0.95\textwidth]
{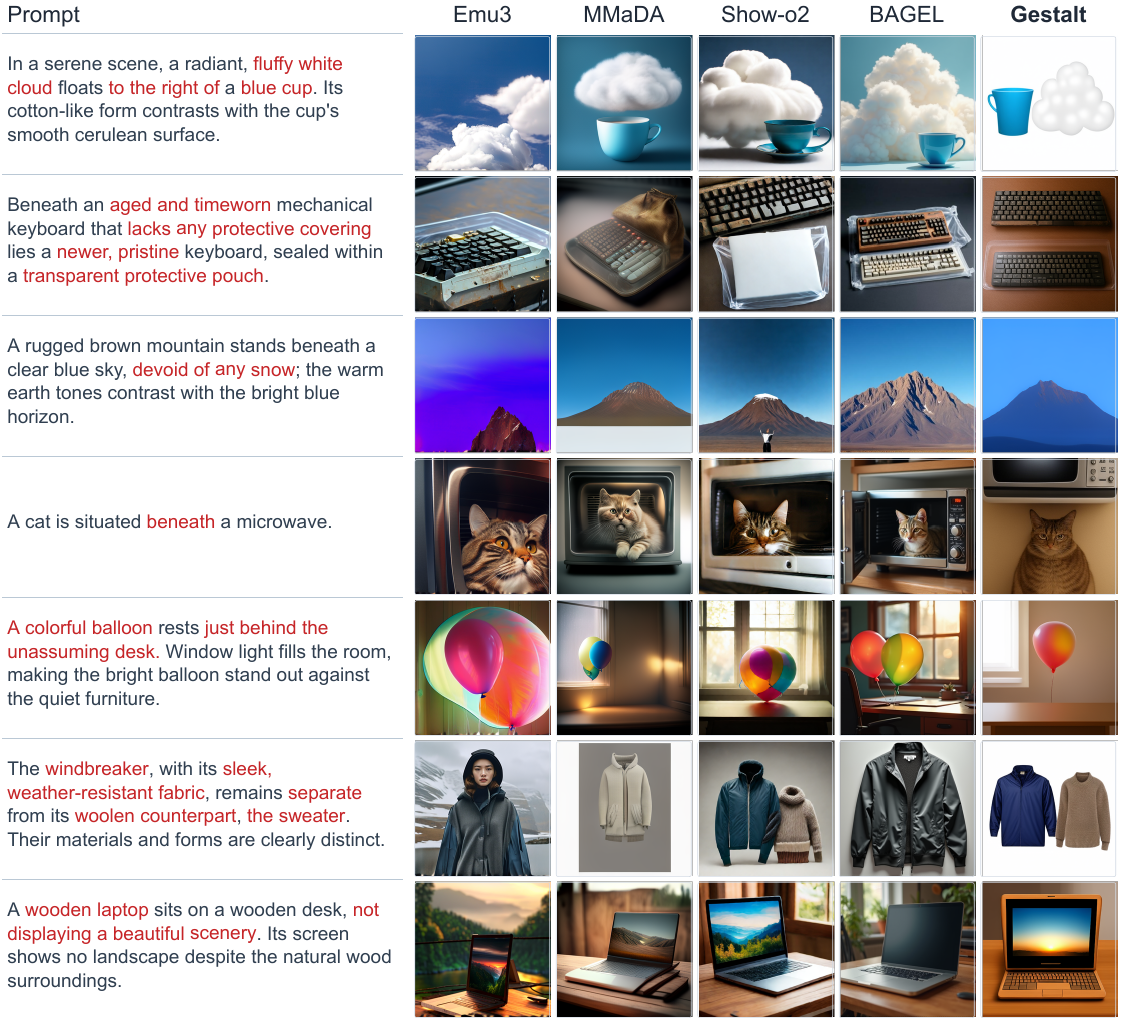}
\caption{
\textbf{Qualitative comparison of fine-grained text-to-image generation.}
\textcolor{red}{Red} text highlights key semantic constraints in the prompts, such as spatial relations, attributes, negation, and properties.
}
\label{fig:model_comparison_qualitative}
\end{figure}

  \begin{table}[t]
\caption{Evaluations on UniGenBench English Long for fine-grained text-to-image generation. The best and the second-best results are highlighted in \textbf{bold} and \underline{underline}, respectively.}
  \label{tab:unigenbench}
  \centering
\newcommand{\centeredscore}[2]{#1\rlap{#2}}
  \resizebox{\textwidth}{!}{
\begin{tabular}{l*{11}{w{c}{34pt}}}
  \specialrule{\heavyrulewidth}{\abovetopsep}{0pt}
  \rowcolor{headercolor}
\rule[-0.5em]{0pt}{1.6em}%
  \textbf{Model} & \textbf{Overall} & \textbf{Style} & \textbf{World} & \textbf{Attr.} & \textbf{Action} & \textbf{Rel.} & \textbf{Comp.} &
  \textbf{Grammar} & \textbf{Layout} & \textbf{Logic} & \textbf{Text} \\
  \specialrule{\lightrulewidth}{0pt}{\belowrulesep}
  \multicolumn{12}{c}{\textit{Gen. Only}} \\
  \midrule
  DALL-E-3
  & 70.82 & \underline{95.08} & \textbf{92.71} &
  \underline{84.98} & 68.36 & 77.90 & 73.88 &
  68.19 & 71.76 & 57.11 & 18.26 \\
  SD-3.5-Large
  & 64.35 & 88.12 & 88.15 & 78.78 & 59.63 & 67.62 &
  62.21 & 65.23 & 71.19 & 44.90 & 17.66 \\
  OmniGen2
  & 71.39 & 94.35 & 84.83 & 83.03 & 66.57 & 73.06 &
  70.49 & \underline{76.40} & 80.63 & 56.55 & \textbf{27.99} \\
  \midrule
  \multicolumn{12}{c}{\textit{Unified}} \\
  \midrule
  Emu3
  & 50.95 & 89.36 & 76.16 & 66.81 & 43.80 & 51.70 &
  46.00 & 50.25 & 56.67 & 27.43 & 1.36 \\
  Show-o2
  & 70.33 & 93.11 & 88.44 & \textbf{86.35} & 69.02 & 77.37 &
  76.45 & 70.30 & 80.63 & 59.71 & 1.90 \\
  Janus-Pro
  & 71.11 & 94.02 & 88.15 & 81.81 & \underline{69.14} &
  \underline{77.96} & \underline{76.53} & 74.62 &
  82.14 & \underline{62.62} & 4.08 \\
  MMaDA
  & 40.10 & 75.83 & 52.75 & 49.90 & 32.42 & 39.06 &
  38.37 & 50.00 & 43.02 & 19.42 & 0.27 \\
  BAGEL
  & 71.26 & 92.44 & \underline{89.31} & 84.21 & 67.62 & 75.70 &
  74.71 & 74.75 & 81.90 & 59.71 & 12.23 \\
  Lumina-DiMOO
  & \underline{71.81} & 86.88 & 88.58 & 83.71 & \textbf{69.66} &
  73.33 & 74.93 & 74.49 & \underline{84.84} &
  58.01 & \underline{23.64} \\
  \midrule
  \rule[-0.45em]{0pt}{1.45em}Gestalt
  & \centeredscore{\textbf{73.30}}{\textcolor{gaincolor}{\fontsize{8}{9.6}\selectfont$\uparrow$\kern-1pt1.49}} & \centeredscore{\textbf{95.85}}{\textcolor{gaincolor}{\fontsize{8}{9.6}\selectfont$\uparrow$\kern-1pt0.77}} & 81.65 & 82.51 & 68.57 &
  \centeredscore{\textbf{78.12}}{\textcolor{gaincolor}{\fontsize{8}{9.6}\selectfont$\uparrow$\kern-1pt0.16}} & \centeredscore{\textbf{81.20}}{\textcolor{gaincolor}{\fontsize{8}{9.6}\selectfont$\uparrow$\kern-1pt4.67}} & \centeredscore{\textbf{82.23}}{\textcolor{gaincolor}{\fontsize{8}{9.6}\selectfont$\uparrow$\kern-1pt5.83}} &
  \centeredscore{\textbf{85.79}}{\textcolor{gaincolor}{\fontsize{8}{9.6}\selectfont$\uparrow$\kern-1pt0.95}} & \centeredscore{\textbf{73.28}}{\textcolor{gaincolor}{\fontsize{8}{9.6}\selectfont$\uparrow$\kern-1pt10.66}} & 3.80 \\
  \bottomrule
  \end{tabular}}
  \end{table}

\begin{table}[t] 
\caption{Evaluations on TIIF-Bench for text-to-image instruction following under short and long prompts. The best and the second-best results are highlighted in \textbf{bold} and \underline{underline}, respectively.} 
\label{tab:tiifbench}
  \centering
\newcommand{\centeredscore}[2]{#1\rlap{#2}}
  \resizebox{\textwidth}{!}{
\begin{tabular}{l*{8}{w{c}{42pt}}@{\hspace{15pt}}}
  \specialrule{\heavyrulewidth}{\abovetopsep}{0pt}
  \rowcolor{headercolor}
\rule[-0.5em]{0pt}{1.6em}%
  \textbf{Model} & \multicolumn{2}{c}{\textbf{Overall}} & \multicolumn{2}{c}{\textbf{Basic}} &
  \multicolumn{2}{c}{\textbf{Advanced}} & \multicolumn{2}{c}{\textbf{Designer}} \\
\noalign{\vskip\dimexpr0.5pt-\aboverulesep\relax}
\cmidrule[\lightrulewidth](lr){2-3}\cmidrule[\lightrulewidth](lr){4-5}
  \cmidrule[\lightrulewidth](lr){6-7}\cmidrule[\lightrulewidth](lr){8-9}
   & \textbf{Short} & \textbf{Long} & \textbf{Short} & \textbf{Long} & \textbf{Short} & \textbf{Long} & \textbf{Short} & \textbf{Long} \\
  \midrule
  \multicolumn{9}{c}{\textit{Gen. Only}} \\
  \midrule
  PixArt-Sigma
  & 62.00 & 58.12 & 70.66 & 75.25 & 57.65 & 49.50 & 62.11 & 52.41 \\
  FLUX.1 Pro
  & 67.32 & 69.89 & 79.08 & 78.91 & 61.10 & 65.37 & 71.80 & 68.80 \\
  MidJourney V7
  & 68.74 & 65.69 & 77.41 & 76.00 & 64.66 & 60.53 & 68.83 & 63.61 \\
  SD 3.5 Large
  & 71.15 & 66.96 & 78.34 & 79.56 & 67.67 & 61.18 & 64.43 & 66.39 \\
  \midrule
  \multicolumn{9}{c}{\textit{Unified}} \\
  \midrule
  Emu3
  & 43.38 & 39.44 & 49.88 & 42.08 & 37.09 & 33.52 & 53.73 & 60.45 \\
  MMaDA
  & 52.32 & 52.90 & 65.78 & 66.71 & 50.32 & 50.76 & 60.45 & 55.60 \\
  Show-o2
  & 67.38 & 68.76 & 81.16 & \underline{84.06} &
    69.25 & \underline{72.99} & \underline{75.37} & \underline{75.75} \\
  Janus-Pro
  & 66.50 & 65.02 & 79.33 & 78.25 & 59.71 & 58.82 & 65.84 & 60.25 \\
  BAGEL
  & \underline{71.50} & \underline{71.70} & \underline{81.79} & 80.05 &
    70.24 & 72.19 & 68.28 & 67.91 \\
  Lumina-DiMOO
  & 71.27 & 68.53& 75.50 & 78.29& \underline{70.49}
 &  68.33 & 69.78& 70.90 \\
  \midrule
  \rule[-0.45em]{0pt}{1.45em}Gestalt
  & \centeredscore{\textbf{74.57}}{\textcolor{gaincolor}{\fontsize{8}{9.6}\selectfont$\uparrow$\kern-1pt3.07}} & \centeredscore{\textbf{78.86}}{\textcolor{gaincolor}{\fontsize{8}{9.6}\selectfont$\uparrow$\kern-1pt7.16}} &
    \centeredscore{\textbf{83.79}}{\textcolor{gaincolor}{\fontsize{8}{9.6}\selectfont$\uparrow$\kern-1pt2.00}} & \centeredscore{\textbf{86.06}}{\textcolor{gaincolor}{\fontsize{8}{9.6}\selectfont$\uparrow$\kern-1pt2.00}} &
    \centeredscore{\textbf{72.60}}{\textcolor{gaincolor}{\fontsize{8}{9.6}\selectfont$\uparrow$\kern-1pt2.11}} & \centeredscore{\textbf{77.69}}{\textcolor{gaincolor}{\fontsize{8}{9.6}\selectfont$\uparrow$\kern-1pt4.70}} &
    \centeredscore{\textbf{84.70}}{\textcolor{gaincolor}{\fontsize{8}{9.6}\selectfont$\uparrow$\kern-1pt9.33}} & \centeredscore{\textbf{88.81}}{\textcolor{gaincolor}{\fontsize{8}{9.6}\selectfont$\uparrow$\kern-1pt13.06}} \\
  \bottomrule
  \end{tabular}}

  \end{table}

\subsection{Unified Multimodal Capabilities}

\subsubsection{Fine-Grained Text-to-Image Generation}

Gestalt demonstrates strong fine-grained text-to-image generation under complex semantic constraints, including multiple objects, attributes, spatial relations, and compositional requirements, as shown in Figure \ref{fig:model_comparison_qualitative}. We compare Gestalt with two groups of baselines: generation-only models specialized for image synthesis, such as DALL-E 3~\citep{betker2023improving} and FLUX.1 Pro~\citep{flux2024}, and unified multimodal models that support both generation and understanding, such as BAGEL~\citep{deng2025emerging} and Janus-Pro~\citep{chen2025janus}.

\noindent\textbf{Fine-grained semantic alignment.}
We evaluate fine-grained text-to-image alignment on UniGenBench~\citep{wang2025unigenbench}, which covers attributes, actions, spatial relations, composition, and logical reasoning. As shown in Table~\ref{tab:unigenbench}, Gestalt achieves the best overall score of $73.30$, surpassing the highest-scoring baselines in the generation-only and unified groups, OmniGen2 ($71.39$) and Lumina-DiMOO ($71.81$), respectively. Gestalt further ranks first in style, relation, composition, grammar, layout, and logic. These dimensions span local semantic details and global compositional structure, requiring the model to preserve relations among concepts and jointly realize multiple constraints. Together, the results reflect precise alignment between textual semantics and visual structure beyond coarse concept matching.

\noindent\textbf{Complex instruction following.}
We further evaluate complex text-to-image instruction following on TIIF-Bench \citep{wei2025tiif}. As Table~\ref{tab:tiifbench}, Gestalt achieves the best overall scores under both short and long instructions, reaching $74.57$ and $78.86$, respectively, ahead of all generation-only and unified baselines. It also ranks first across all three evaluation categories in both settings. Its advantage grows under longer instructions, which introduce more entities and interdependent constraints, demonstrating stronger retention and joint realization of detailed semantic requirements. Together, UniGenBench and TIIF-Bench show that Gestalt unifies understanding and generation without sacrificing fine-grained text-to-image performance, outperforming even specialized generation models.

\begin{table}[t]
\centering
\newcommand{\centeredscore}[2]{#1\rlap{#2}}
\caption{Evaluations on multimodal understanding benchmarks. The symbol $^\dagger$ denotes the results from the official paper, while the rest of the results are evaluated using the official checkpoint and inference scripts. The \textbf{best} and the \underline{second-best} results in diffusion-based models are highlighted.}
\label{tab:diffusion_ar_comparison}
\fontsize{8.5}{10.2}\selectfont
\setlength{\tabcolsep}{3.48pt}
\begin{tabular}{l*{8}{w{c}{34pt}}@{\hspace{7pt}}}
\specialrule{\heavyrulewidth}{\abovetopsep}{0pt}
\rowcolor{headercolor}
\rule[-0.5em]{0pt}{1.6em}%
\textbf{Models}
                         & \multicolumn{5}{c}{\textbf{General}}
                         & \multicolumn{3}{c}{\textbf{Vision-Centric}} \\
\noalign{\vskip\dimexpr0.5pt-\aboverulesep\relax}
\cmidrule[\lightrulewidth](lr){2-6} \cmidrule[\lightrulewidth](lr){7-9}
                         & \textbf{MME-P}
                         & \textbf{GQA}
                         & \textbf{MMStar-P}
                         & \textbf{POPE}
                         & \textbf{RWQA}
                         & \textbf{MMVP}
                         & \textbf{CVB}$^\mathrm{2d}$
                         & \textbf{CVB}$^\mathrm{3d}$ \\
\midrule

                         & \multicolumn{8}{c}{\textit{AR-Based}} \\
\midrule

BAGEL
                         & \centeredscore{1687.0}{$^\dagger$} 
                         & 66.4
                         & 70.9
                         & 88.2
                         & 67.6
                         & \centeredscore{69.3}{$^\dagger$}
                         & 77.7
                         & 84.2 \\
\midrule

                         & \multicolumn{8}{c}{\textit{Diffusion-based}} \\
\midrule

MMaDA
                         & \centeredscore{1410.7}{$^\dagger$}
                         & \centeredscore{\textbf{61.3}}{$^\dagger$}
                         & 43.0
                         & \centeredscore{86.1}{$^\dagger$}
                         & 48.2
                         & 17.3
                         & 55.3
                         & 54.8 \\

Lumina-DiMOO
                         & \centeredscore{\underline{1534.2}}{$^\dagger$}
                         & 43.3
                         & -
                         & \centeredscore{\textbf{87.4}}{$^\dagger$}
                         & 35.9
                         & 34.0
                         & 54.3
                         & 52.0 \\


LaViDA-o
                         & 1431.0
                         & 54.1
                         & \underline{55.9}
                         & --
                         & 56.6
                         & \underline{47.3}
                         & 73.4
                         & 70.8 \\

LLaDA-o
                         & \centeredscore{1412.0}{$^\dagger$}
                         & 58.0
                         & 55.6
                         & \underline{87.2}
                         & \textbf{66.4}
                         & 46.7
                         & \underline{78.0}
                         & \underline{75.9} \\

Omni-diffusion
                         & \centeredscore{1216.7}{$^\dagger$}
                         & --
                         & --
                         & \centeredscore{76.6}{$^\dagger$}
                         & --
                         & --
                         & --
                         & -- \\

\midrule
\rule[-0.45em]{0pt}{1.45em}Gestalt
                         & \centeredscore{\textbf{1600.2}}{\textcolor{gaincolor}{\fontsize{6.8}{8.16}\selectfont$\uparrow$\kern-1pt66.0}}
                         & \underline{60.2}
                         & \textbf{60.8}
                         & 86.5
                         & \underline{59.1}
                         & \centeredscore{\textbf{48.7}}{\textcolor{gaincolor}{\fontsize{6.8}{8.16}\selectfont$\uparrow$\kern-1pt1.4}}
                         & \centeredscore{\textbf{78.9}}{\textcolor{gaincolor}{\fontsize{6.8}{8.16}\selectfont$\uparrow$\kern-1pt0.9}}
                         & \centeredscore{\textbf{86.1}}{\textcolor{gaincolor}{\fontsize{6.8}{8.16}\selectfont$\uparrow$\kern-1pt10.2}} \\

\bottomrule
\end{tabular}
\end{table}

\subsubsection{Multimodal Understanding}

Gestalt is evaluated on eight multimodal understanding benchmarks, grouped into (1) general VQA benchmarks: MME-P~\citep{fu2026mme}, GQA~\citep{hudson2019gqa}, MMStar-Perception~\citep{chen2024we}, POPE~\citep{li2023evaluating}, and RealworldQA~\citep{xai2024grok15v}; and (2) vision-centric benchmarks: MMVP~\citep{tong2024eyes} and CV-Bench~\citep{tong2024cambrian1}. We compare Gestalt with diffusion-based unified multimodal models, including MMaDA~\citep{yang2026mmada}, Lumina-DiMOO~\citep{xin2025lumina}, Lavida-o~\citep{li2026lavida}, LLaDA-o~\citep{you2026llada}, and Omni-diffusion~\citep{li2026omni}, as well as BAGEL~\citep{deng2025emerging}, a representative autoregressive unified model. As shown in Table~\ref{tab:diffusion_ar_comparison}, Gestalt achieves the best results among diffusion-based models on five benchmarks and ranks among the top two on seven of the eight benchmarks. This performance establishes Gestalt as a strong diffusion-based unified model across both general multimodal understanding and vision-centric perception.
Gestalt shows particularly strong performance on vision-centric understanding. It achieves the best results among diffusion-based models on MMVP and both CV-Bench tracks. Notably, Gestalt also surpasses BAGEL, a representative autoregressive unified model, on both CV-Bench tracks: Gestalt scores $78.9$ and $86.1$, whereas the AR-based BAGEL obtains $77.7$ and $84.2$, respectively. These gains on vision-centric benchmarks align with Gestalt's explicit emphasis on modality-unique information and highlight the potential of interplay-oriented modeling to strengthen visual understanding within a unified diffusion model.

\begin{figure}[t]
\centering
\includegraphics[width=0.9\linewidth]{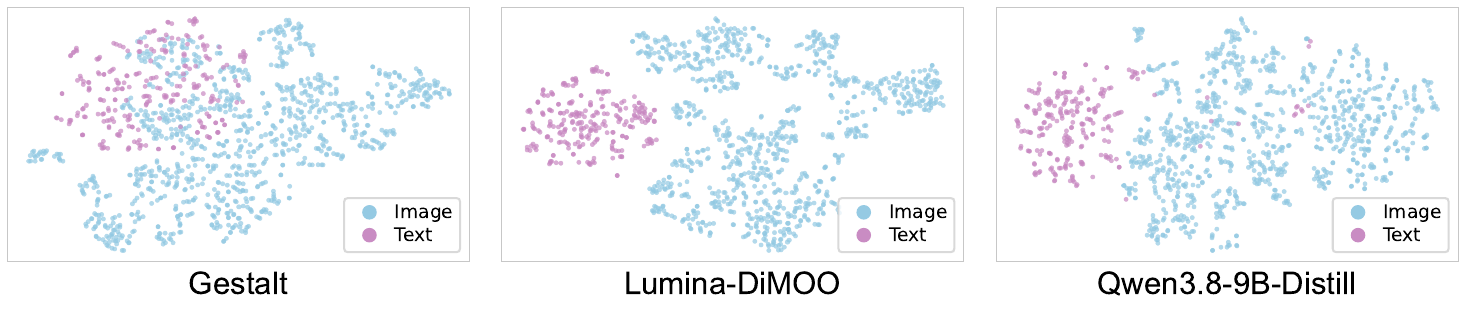}
\caption{Visualization of image-text hidden representation across diffusion and AR architecture.}
\label{embedding_comparison}

\end{figure}

\subsection{Cross-modal Representation Analysis}

Beyond benchmark performance, we examine the representations underlying Gestalt’s multimodal capabilities, including the organization of visual and textual representations and the preservation of pretrained language representations and text-only capability.


\noindent\textbf{Unified Vision-Language Representation Space.}
We examine how visual and textual token representations are organized in the representation space at layer 24 using image--caption pairs from the DOCCI dataset~\citep{onoe2024docci}. We compare Gestalt with Lumina-DiMOO~\citep{xin2025lumina}, a diffusion-based unified model, and Qwen3.8-9B-Distill\footnote{As no official 9B checkpoint of Qwen3.8 is publicly available, we use the distilled variant Qwen3.8-9B-Distill: \url{https://huggingface.co/empero-ai/Qwen3.8-9B-Distill}.}, an autoregressive native multimodal model, pretrained from scratch with early-fused multimodal data. As shown in \autoref{embedding_comparison}, Lumina-DiMOO exhibits a clear separation between the two modalities, while Qwen3.8 shows partial cross-modal overlap, with a substantial portion of textual tokens remaining separated from the visual representations. In contrast, Gestalt exhibits a more interleaved distribution of visual and textual tokens, suggesting a more integrated vision-language representation space than both diffusion-based and autoregressive baselines.


\begin{table}[t]
    \vspace{-1em}
    \centering
    \setlength{\tabcolsep}{4pt}
    \caption{
    Comparison of distribution shift (Norm and Distance) and representation preservation (Cosine Similarity and CKA) of LLaDA-based multimodal models in the language embedding space.
    }

    \label{tab:embedding_norm_comparison}
    \fontsize{8.5}{10.2}\selectfont
    \begin{tabular}{llccccc}
        \specialrule{\heavyrulewidth}{\abovetopsep}{0pt}
        \rowcolor{headercolor}
\rule[-0.5em]{0pt}{1.6em}%
        \textbf{Type}
        & \textbf{Model}
        & \textbf{Norm Mean}
        & \textbf{Norm Std}
        & \textbf{Mean Distance}
        & \textbf{Cosine Sim.}
        & \textbf{CKA}
        \\
        \specialrule{\lightrulewidth}{0pt}{\belowrulesep}

        Base
        & LLaDA
        & 7.82 & 1.11 & -- & -- & -- \\

        \midrule

        \multirow{3}{*}{Projector Alignment}
        & LLaDA-V
        & 7.82 & 1.11
        & 0.043
        & 0.99998
        & 0.99997
        \\

        & LaViDa
        & 7.83 & 1.11
        & 0.160
        & 0.99976
        & 0.99964
        \\

        & LLaDA-O
        & 8.02 & 1.16
        & 1.520
        & 0.98136
        & 0.97424
        \\

        \midrule

        \multirow{2}{*}{Unified Token Space}
        & Lumina-DiMOO
        & 0.85 & 0.49
        & 7.426
        & 0.58795
        & 0.43894
        \\

        & Gestalt
        & 5.29 & 0.76
        & 2.590
        & 0.99677
        & 0.99494
        \\

        \bottomrule
    \end{tabular}

\end{table}

\noindent\textbf{Representation Preservation under Multimodal Adaptation.}
We compare how two multimodal adaptation paradigms reshape the language representation space. Projector-alignment models align visual features to the language space while preserving textual representations, whereas unified token-space models jointly represent vision and language within a shared discrete token space. We report representation changes using the mean and standard deviation of embedding norms, and measure preservation relative to pretrained LLaDA using distance, cosine similarity, and CKA.
As shown in Table~\ref{tab:embedding_norm_comparison}, projector-alignment models introduce only minor changes to the original language space, while unified models exhibit more substantial distribution shifts. Notably, Lumina-DiMOO reduces cosine similarity and CKA with pretrained LLaDA to $0.5880$ and $0.4389$, respectively, whereas Gestalt maintains substantially higher values of $0.9968$ and $0.9949$ despite a noticeable shift in embedding scale. These results suggest that Gestalt build a new space for unified multimodal modeling while largely preserving its original representation structure.

\begin{table}[t]
\caption{Evaluation of language capabilities based on six text-only benchmarks. The \textbf{best} and the \underline{second-best} results among diffusion-based unified models are highlighted.}
\label{tab:text_only}
\centering
\newcommand{\centeredscore}[2]{#1\rlap{#2}}
\fontsize{8.5}{10.2}\selectfont
\setlength{\tabcolsep}{4pt}
\begin{tabular}{l*{6}{w{c}{44pt}}@{\hspace{9pt}}}
\specialrule{\heavyrulewidth}{\abovetopsep}{0pt}
\rowcolor{headercolor}[\tabcolsep][9pt]
\rule[-0.5em]{0pt}{1.6em}%
\textbf{Model} & \textbf{MMLU} & \textbf{TruthfulQA} & \textbf{WinoGrande} & \textbf{HellaSwag} & \textbf{ARC-E} & \textbf{ARC-C} \\
\specialrule{\lightrulewidth}{0pt}{\belowrulesep}
\multicolumn{7}{c}{\textit{AR-Based}} \\
\midrule
Show-o2 & 71.70 & 46.94 & 74.03 & 76.67 & 84.43 & 58.53 \\
Janus-Pro & 49.90 & 41.72 & 67.17 & 68.41 & 65.74 & 40.70 \\
BAGEL & 28.02 & 40.51 & 50.75 & 28.59 & 27.53 & 23.63 \\
\midrule
\multicolumn{7}{c}{\textit{Diffusion-based}} \\
\midrule
MMaDA & \underline{40.14} & 43.81 & \underline{54.85} & \underline{45.81} & \underline{46.72} & \underline{28.67} \\
Lumina-DiMOO & 29.75 & 43.34 & 51.62 & 39.11 & 44.44 & 26.45 \\
LLaDA-o & 25.25 & \textbf{52.30} & 51.22 & 30.72 & 36.66 & 24.66 \\
\midrule
\rule[-0.45em]{0pt}{1.45em}Gestalt & \centeredscore{\textbf{49.50}}{\textcolor{gaincolor}{\fontsize{6.8}{8.16}\selectfont$\uparrow$\kern-1pt9.36}} & \underline{47.55} & \centeredscore{\textbf{60.62}}{\textcolor{gaincolor}{\fontsize{6.8}{8.16}\selectfont$\uparrow$\kern-1pt5.77}} & \centeredscore{\textbf{53.44}}{\textcolor{gaincolor}{\fontsize{6.8}{8.16}\selectfont$\uparrow$\kern-1pt7.63}} & \centeredscore{\textbf{62.29}}{\textcolor{gaincolor}{\fontsize{6.8}{8.16}\selectfont$\uparrow$\kern-1pt15.57}} & \centeredscore{\textbf{43.69}}{\textcolor{gaincolor}{\fontsize{6.8}{8.16}\selectfont$\uparrow$\kern-1pt15.02}} \\
\bottomrule
\end{tabular}
\vspace{-1em}
\end{table}

\noindent\textbf{Language Capability.}
We next assess representation-level preservation through text-only capability. We compare Gestalt with diffusion-based unified models MMaDA, Lumina-DiMOO, and LLaDA-o, and AR-based unified models Show-o2, Janus-Pro, and BAGEL~\citep{yang2026mmada,xin2025lumina,you2026llada,xie2026show,chen2025janus,deng2025emerging}. Evaluation is 5-shot on MMLU, WinoGrande, and TruthfulQA and 0-shot on HellaSwag, ARC-Easy, and ARC-Challenge, with TruthfulQA reported using MC2. As shown in Table~\ref{tab:text_only}, Gestalt leads diffusion-based unified models on five of six benchmarks, with particularly large gains on ARC-E and ARC-C. Notably, it is also competitive with AR-based unified models, outperforming BAGEL across all six benchmarks, closely matching Janus-Pro on MMLU, and surpassing it on TruthfulQA and ARC-C. These results confirm that Gestalt retains strong language capability after multimodal adaptation.

\subsection{Understanding the Role of Cross-Modal Interplay}
Multimodal interplay is central to Gestalt, preserving modality-unique information while deriving synergy through cross-modal integration. This section evaluates these capabilities and examines how interplay tokens encode their distinct information demands.

\begin{wraptable}{r}{0.46\columnwidth}
\centering
\caption{
Evaluation on multimodal benchmarks requiring visual-specific and synergistic interplay.
}
\label{tab:interplay_eval}

\fontsize{6.8}{8.0}\selectfont
\setlength{\tabcolsep}{1.5pt}
\renewcommand{\arraystretch}{1.0}

\newcommand{\gain}[1]{%
  \hspace{0.4pt}%
  {\fontsize{5.3}{6.2}\selectfont
   \textcolor{gaincolor}{$\uparrow$#1}}%
}

\newcolumntype{L}[1]{%
  >{\hsize=#1\hsize\linewidth=\hsize
    \raggedright\arraybackslash}X%
}
\newcolumntype{C}[1]{%
  >{\hsize=#1\hsize\linewidth=\hsize
    \centering\arraybackslash}X%
}

\begin{tabularx}{\linewidth}{
  @{}
  L{1.35}
  C{0.75}
  C{1.20}
  C{0.90}
  C{0.80}
  @{}
}

\specialrule{\heavyrulewidth}{\abovetopsep}{0pt}

\rowcolor{headercolor}
\rule[-0.2em]{0pt}{1.4em}
{\bfseries Model}
& \multicolumn{2}{c}{{\bfseries Visual}}
& \multicolumn{2}{c}{{\bfseries Synergy}} \\

\rule[-0.4em]{0pt}{1.5em}
& {\bfseries\mbox{MIB-V}}
& {\bfseries\mbox{CoreCog-SM}}
& {\bfseries\mbox{MM-IMDb}}
& {\bfseries\mbox{SRBench}} \\

\specialrule{\lightrulewidth}{0pt}{\belowrulesep}

\multicolumn{5}{c}{\textit{AR-Based}} \\
\midrule

BAGEL
& 65.96
& 65.00
& 60.60
& 51.89 \\

\midrule

\multicolumn{5}{c}{\textit{Diffusion-Based}} \\
\midrule

MMaDA
& 44.23
& 43.20
& 30.39
& 36.72 \\

Lumina-DiMOO
& 56.43
& 42.20
& 31.22
& 45.50 \\

LLaDA-o
& {\bfseries 60.31}
& \underline{53.30}
& 39.64
& \underline{50.89} \\

LaViDa-O
& 58.07
& 50.60
& \underline{51.54}
& 40.89 \\

\midrule

Gestalt
& \underline{59.89}
& {\bfseries 55.20}\gain{1.90}
& {\bfseries 67.07}\gain{15.53}
& {\bfseries 53.17}\gain{2.28} \\

\bottomrule
\end{tabularx}
\vspace{-1em}

\end{wraptable}

\noindent\textbf{Evaluation of Multimodal Interplay.}
We evaluate two representative forms of multimodal interplay. MIBench-Visual subset (MIB-V)~\citep{miao2026mibench} and CoreCognition Sensorimotor subset (CoreCog-SM)~\citep{li2024core} are designed for evaluating basic visual-specific capabilities, while MM-IMDb~\citep{arevalo2017gated} and SRBench~\citep{stogiannidis2025mind} focus on cross-modal synergy through multimodal integration and reasoning. As shown in Table~\ref{tab:interplay_eval}, Gestalt achieves strong performance on both visual-oriented and synergy-oriented benchmarks, including the best results among diffusion-based models on CoreCognition, MM-IMDb, and SRBench. These results demonstrate that Gestalt can effectively handle different forms of multimodal interplay, maintaining strong visual capability while benefiting from cross-modal synergy.

\newpage
\begin{wrapfigure}{r}{0.3\textwidth}
\centering
\vspace{-1em}
\includegraphics[width=1\linewidth]{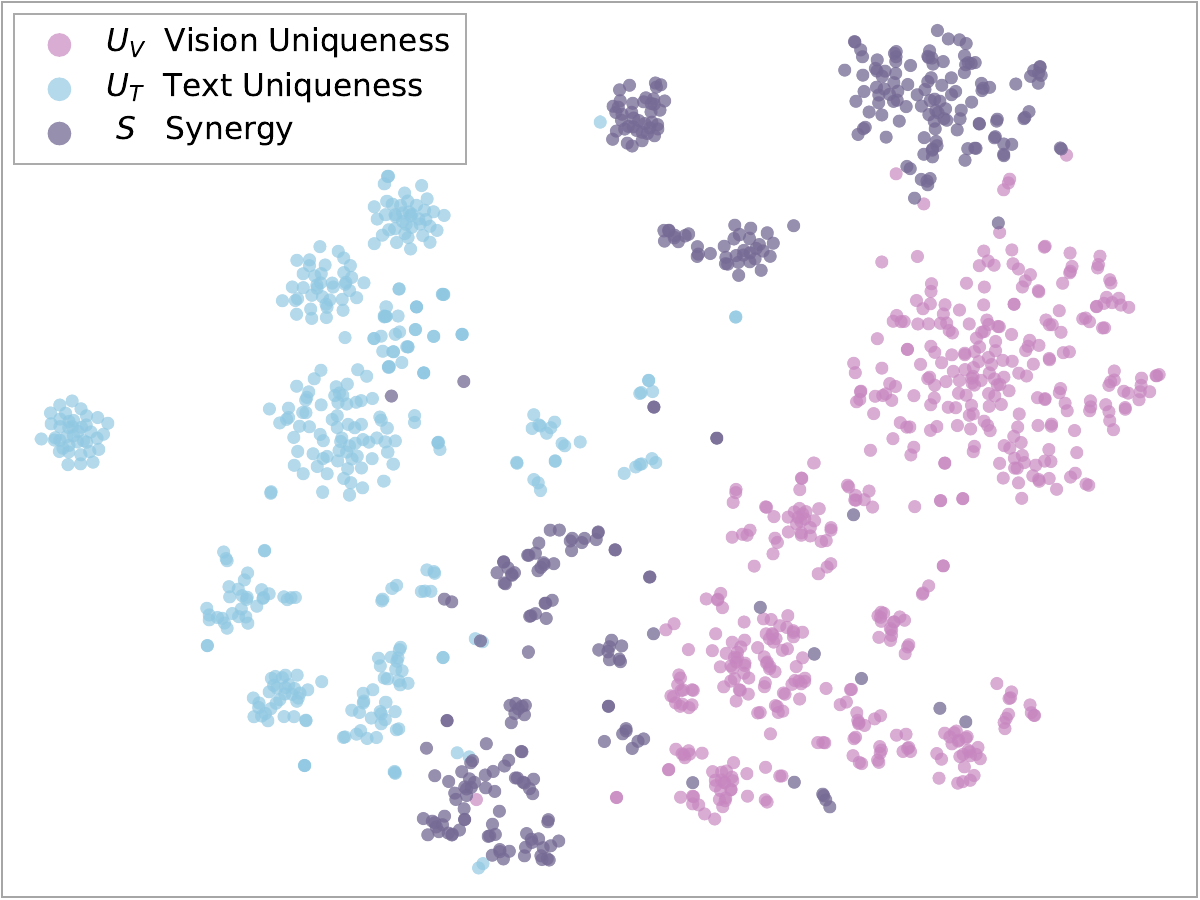}
\caption{Visualization of interplay token distribution across samples with different multimodal interplay.}
\label{fig:interplay_distribution}
\end{wrapfigure}

\noindent\textbf{Interplay-Aware Representations.}
We further analyze how interplay tokens respond to samples with different forms of multimodal interplay. Specifically, we visualize their hidden representations at layer 16 on MIBench, which categorizes samples into visual uniqueness, textual uniqueness, and cross-modal synergy according to the information required for prediction. As shown in Figure~\ref{fig:interplay_distribution}, the interplay-token representations exhibit clear structures associated with different interplay types. Samples dominated by visual or textual information occupy largely separated regions, whereas synergy samples form a distinct distribution that partially bridges the two. These patterns suggest that interplay tokens adapt to different multimodal information requirements, reflecting whether prediction depends primarily on visual information, textual information, or information that emerges from their combination.

\section{Related Work}
Large multimodal models have evolved from architectural compatibility to data compatibility. Early systems connect modality-specific vision encoders to pretrained AR-based language models through learned projectors~\citep{liu2023visual,bai2025qwen25vltechnicalreport}, whereas native unified multimodal models jointly pretrain visual and textual data in a shared model to support both understanding and generation~\citep{zhao2025unified}. These unified models use AR next-token prediction~\citep{team2024chameleon,wang2024emu3,chen2025janus}, discrete diffusion~\citep{nie2026large,you2026llada,yu2025dimple,li2026lavida,yang2026mmada,xin2025lumina}, or hybrid AR-diffusion and flow-based formulations~\citep{zhou2025transfusion,xie2025show,xie2026show}. Despite this progress, existing work largely focuses on bringing visual and textual data into a shared model, without explicitly organizing the information relationships and emergent properties that arise across modalities.

\section{Discussion}
Moving beyond simply accommodating additional modalities, Gestalt frames unified multimodal modeling around a multistage interplay pyramid. Our findings indicate that unification need not require sharing all computation from the earliest layer; modality-specific processing and controlled exchange before full integration may better balance specialization and cross-modal synergy. In existing studies of mixed-modal scaling laws, multimodal synergy is treated as a key component~\citep{aghajanyan2023scaling}. This connection suggests that multimodal scaling may depend not only on model and data size, but also on how information exchange is organized across stages. The interplay pyramid provides a structured setting for examining this possibility and may therefore offer a useful basis for future research on scaling laws in large multimodal models.

\section*{Acknowledgments}
This work is supported in part by the Beijing Natural Science Foundation under Grant No.~4262050 and by the Beijing Nova Program under Grant No.~202604841277. 

\noindent \textbf{Author Contributions. } Zequn Yang, Yake Wei, and Di Hu drove the overall advancement of the project. Zequn Yang, Yu Miao, Haotian Ni, Ziheng Chen, and Chengxiang Huang contributed equally to this work. Zequn Yang conducted model pretraining and supervised fine-tuning. Zequn Yang, Yu Miao, and Haotian Ni developed the model architecture and conducted the core experiments. Yu Miao, Ziheng Chen, and Chengxiang Huang contributed to data processing and organization. Yu Miao conducted the evaluation of image generation capabilities, Haotian Ni conducted the multimodal understanding evaluation, and Ziheng Chen conducted the text-only evaluation. Dongzhan Zhou, Kai Chen, Qi Zhang, and Ji-Rong Wen contributed to discussions on the technical design and methodology. Yake Wei, Zequn Yang, Yu Miao, Haotian Ni, Ziheng Chen, and Di Hu contributed to writing and revising the manuscript. Di Hu initiated the project. Yake Wei and Di Hu supervised and advised the project.



\clearpage

\bibliographystyle{IEEEtran}
\bibliography{paper}

@article{liu2023visual,
  title={Visual instruction tuning},
  author={Liu, Haotian and Li, Chunyuan and Wu, Qingyang and Lee, Yong Jae},
  journal={Advances in neural information processing systems},
  volume={36},
  pages={34892--34916},
  year={2023}
}

@article{team2024chameleon,
  title={Chameleon: Mixed-modal early-fusion foundation models},
  author={Team, Chameleon},
  journal={arXiv preprint arXiv:2405.09818},
  year={2024}
}

@article{nie2026large,
  title={Large language diffusion models},
  author={Nie, Shen and Zhu, Fengqi and You, Zebin and Zhang, Xiaolu and Ou, Jingyang and Hu, Jun and Zhou, Jun and Lin, Yankai and Wen, Ji-Rong and Li, Chongxuan},
  journal={Advances in Neural Information Processing Systems},
  volume={38},
  pages={50608--50646},
  year={2026}
}

@inproceedings{you2026llada,
  title={Llada-v: Large language diffusion models with visual instruction tuning},
  author={You, Zebin and Nie, Shen and Zhang, Xiaolu and ZHOU, JUN and Lu, Zhiwu and Wen, Ji-Rong and Li, Chongxuan},
  booktitle={Proceedings of the IEEE/CVF Conference on Computer Vision and Pattern Recognition},
  pages={10093--10105},
  year={2026}
}

@article{yu2025dimple,
  title={Dimple: Discrete diffusion multimodal large language model with parallel decoding},
  author={Yu, Runpeng and Ma, Xinyin and Wang, Xinchao},
  journal={arXiv preprint arXiv:2505.16990},
  year={2025}
}

@article{zhao2025unified,
  title={Unified multimodal understanding and generation models: Advances, challenges, and opportunities},
  author={Zhao, Shanshan and Zhang, Xinjie and Guo, Jintao and Hu, Jiakui and Duan, Lunhao and Fu, Minghao and Chng, Yong Xien and Wang, Guo-Hua and Chen, Qing-Guo and Xu, Zhao and others},
  journal={arXiv preprint arXiv:2505.02567},
  year={2025}
}

@inproceedings{zhou2025transfusion,
  title={Transfusion: Predict the next token and diffuse images with one multi-modal model},
  author={Zhou, Chunting and Yu, Lili and Babu, Arun and Tirumala, Kushal and Yasunaga, Michihiro and Shamis, Leonid and Kahn, Jacob and Ma, Xuezhe and Zettlemoyer, Luke and Levy, Omer},
  booktitle={International Conference on Learning Representations},
  volume={2025},
  pages={6446--6469},
  year={2025}
}

@inproceedings{xie2025show,
  title={Show-o: One single transformer to unify multimodal understanding and generation},
  author={Xie, Jinheng and Mao, Weijia and Bai, Zechen and Zhang, David Junhao and Wang, Weihao and Lin, Kevin Qinghong and Gu, Yuchao and Chen, Zhijie and Yang, Zhenheng and Shou, Mike Zheng},
  booktitle={International Conference on Learning Representations},
  volume={2025},
  pages={28240--28264},
  year={2025}
}

@article{xie2026show,
  title={Show-o2: Improved native unified multimodal models},
  author={Xie, Jinheng and Yang, Zhenheng and Shou, Mike Zheng},
  journal={Advances in Neural Information Processing Systems},
  volume={38},
  pages={47490--47518},
  year={2026}
}

@article{wang2024emu3,
  title={Emu3: Next-token prediction is all you need},
  author={Wang, Xinlong and Zhang, Xiaosong and Luo, Zhengxiong and Sun, Quan and Cui, Yufeng and Wang, Jinsheng and Zhang, Fan and Wang, Yueze and Li, Zhen and Yu, Qiying and others},
  journal={arXiv preprint arXiv:2409.18869},
  year={2024}
}

@article{yang2026mmada,
  title={Mmada: Multimodal large diffusion language models},
  author={Yang, Ling and Tian, Ye and Li, Bowen and Zhang, Xinchen and Shen, Ke and Tong, Yunhai and Wang, Mengdi},
  journal={Advances in Neural Information Processing Systems},
  volume={38},
  pages={138867--138907},
  year={2026}
}

@article{xin2025lumina,
  title={Lumina-dimoo: An omni diffusion large language model for multi-modal generation and understanding},
  author={Xin, Yi and Qin, Qi and Luo, Siqi and Zhu, Kaiwen and Yan, Juncheng and Tai, Yan and Lei, Jiayi and Cao, Yuewen and Wang, Keqi and Wang, Yibin and others},
  journal={arXiv preprint arXiv:2510.06308},
  year={2025}
}

@article{deng2025emerging,
  title={Emerging properties in unified multimodal pretraining},
  author={Deng, Chaorui and Zhu, Deyao and Li, Kunchang and Gou, Chenhui and Li, Feng and Wang, Zeyu and Zhong, Shu and Yu, Weihao and Nie, Xiaonan and Song, Ziang and others},
  journal={arXiv preprint arXiv:2505.14683},
  year={2025}
}

@article{li2026lavida,
  title={Lavida: A large diffusion language model for multimodal understanding},
  author={Li, Shufan and Kallidromitis, Konstantinos and Bansal, Hritik and Gokul, Akash and Kato, Yusuke and Kozuka, Kazuki and Kuen, Jason and Lin, Zhe and Chang, Kai-Wei and Grover, Aditya},
  journal={Advances in Neural Information Processing Systems},
  volume={38},
  pages={105101--105134},
  year={2026}
}

@article{li2026omni,
  title={Omni-diffusion: Unified multimodal understanding and generation with masked discrete diffusion},
  author={Li, Lijiang and Long, Zuwei and Shen, Yunhang and Gao, Heting and Cao, Haoyu and Sun, Xing and Shan, Caifeng and He, Ran and Fu, Chaoyou},
  journal={arXiv preprint arXiv:2603.06577},
  year={2026}
}

@article{wang2025unigenbench,
  title={UniGenBench++: A Unified Semantic Evaluation Benchmark for Text-to-Image Generation},
  author={Wang, Yibin and Li, Zhimin and Zang, Yuhang and Bu, Jiazi and Zhou, Yujie and Xin, Yi and He, Junjun and Wang, Chunyu and Lu, Qinglin and Jin, Cheng and others},
  journal={arXiv preprint arXiv:2510.18701},
  year={2025}
}

@article{wei2025tiif,
  title={TIIF-Bench: How Does Your T2I Model Follow Your Instructions?},
  author={Wei, Xinyu and Zhang, Jinrui and Wang, Zeqing and Wei, Hongyang and Guo, Zhen and Zhang, Lei},
  journal={arXiv preprint arXiv:2506.02161},
  year={2025}
}

@article{chen2025janus,
  title={Janus-pro: Unified multimodal understanding and generation with data and model scaling},
  author={Chen, Xiaokang and Wu, Zhiyu and Liu, Xingchao and Pan, Zizheng and Liu, Wen and Xie, Zhenda and Yu, Xingkai and Ruan, Chong},
  journal={arXiv preprint arXiv:2501.17811},
  year={2025}
}

@inproceedings{fu2026mme,
  title   = {{MME}: A Comprehensive Evaluation Benchmark for Multimodal Large Language Models},
  author  = {Fu, Chaoyou and Chen, Peixian and Shen, Yunhang and Qin, Yulei and Zhang, Mengdan and Lin, Xu and Yang, Jinrui and Zheng, Xiawu and Li, Ke and Sun, Xing and others},
  booktitle = {Advances in Neural Information Processing Systems},
  volume  = {38},
  year    = {2026}
}

@inproceedings{hudson2019gqa,
  title     = {{GQA}: A New Dataset for Real-World Visual Reasoning and Compositional Question Answering},
  author    = {Hudson, Drew A. and Manning, Christopher D.},
  booktitle = {Proceedings of the IEEE/CVF Conference on Computer Vision and Pattern Recognition},
  pages     = {6700--6709},
  year      = {2019}
}

@inproceedings{li2023evaluating,
  title={Evaluating object hallucination in large vision-language models},
  author={Li, Yifan and Du, Yifan and Zhou, Kun and Wang, Jinpeng and Zhao, Xin and Wen, Ji-Rong},
  booktitle={Proceedings of the 2023 conference on empirical methods in natural language processing},
  pages={292--305},
  year={2023}
}

@misc{xai2024grok15v,
  title        = {Grok-1.5 Vision Preview},
  author       = {{xAI}},
  howpublished = {\url{https://x.ai/news/grok-1.5v}},
  year         = {2024},
  month        = apr,
  day          = {12},
  note         = {Accessed: 2026-05-20}
}

@inproceedings{tong2024eyes,
  title     = {Eyes Wide Shut? Exploring the Visual Shortcomings of Multimodal {LLMs}},
  author    = {Tong, Shengbang and Liu, Zhuang and Zhai, Yuexiang and Ma, Yi and LeCun, Yann and Xie, Saining},
  booktitle = {Proceedings of the IEEE/CVF Conference on Computer Vision and Pattern Recognition},
  pages     = {9568--9578},
  year      = {2024}
}

@misc{tong2024cambrian1,
      title={Cambrian-1: A Fully Open, Vision-Centric Exploration of Multimodal LLMs},
      author={Shengbang Tong and Ellis Brown and Penghao Wu and Sanghyun Woo and Manoj Middepogu and Sai Charitha Akula and Jihan Yang and Shusheng Yang and Adithya Iyer and Xichen Pan and Austin Wang and Rob Fergus and Yann LeCun and Saining Xie},
      year={2024},
      eprint={2406.16860},
}

@article{tang2016interactions,
  title={The interactions of multisensory integration with endogenous and exogenous attention},
  author={Tang, Xiaoyu and Wu, Jinglong and Shen, Yong},
  journal={Neuroscience \& Biobehavioral Reviews},
  volume={61},
  pages={208--224},
  year={2016},
  publisher={Elsevier}
}

@misc{bai2025qwen25vltechnicalreport,
      title={Qwen2.5-VL Technical Report}, 
      author={Shuai Bai and Keqin Chen and Xuejing Liu and Jialin Wang and Wenbin Ge and Sibo Song and Kai Dang and Peng Wang and Shijie Wang and Jun Tang and Humen Zhong and Yuanzhi Zhu and Mingkun Yang and Zhaohai Li and Jianqiang Wan and Pengfei Wang and Wei Ding and Zheren Fu and Yiheng Xu and Jiabo Ye and Xi Zhang and Tianbao Xie and Zesen Cheng and Hang Zhang and Zhibo Yang and Haiyang Xu and Junyang Lin},
      year={2025},
      eprint={2502.13923},
      archivePrefix={arXiv},
      primaryClass={cs.CV},
      url={https://arxiv.org/abs/2502.13923}, 
}

@misc{kimiteam2026kimik3openfrontier,
  title={{Kimi K3}: Open Frontier Intelligence},
  author={{Kimi Team} and others},
  year={2026},
  eprint={2607.24653},
  archivePrefix={arXiv},
  primaryClass={cs.CL},
  url={https://arxiv.org/abs/2607.24653}
}

@article{kohler1967gestalt,
  title={Gestalt psychology},
  author={K{\"o}hler, Wolfgang},
  journal={Psychologische forschung},
  volume={31},
  number={1},
  pages={XVIII--XXX},
  year={1967},
  publisher={Springer}
}

@inproceedings{shi2025scalable,
  title={Scalable image tokenization with index backpropagation quantization},
  author={Shi, Fengyuan and Luo, Zhuoyan and Ge, Yixiao and Yang, Yujiu and Shan, Ying and Wang, Limin},
  booktitle={2025 IEEE/CVF International Conference on Computer Vision (ICCV)},
  pages={16037--16046},
  year={2025},
  organization={IEEE}
}

@article{an2025llava,
  title={Llava-onevision-1.5: Fully open framework for democratized multimodal training},
  author={An, Xiang and Xie, Yin and Yang, Kaicheng and Zhang, Wenkang and Zhao, Xiuwei and Cheng, Zheng and Wang, Yirui and Xu, Songcen and Chen, Changrui and Zhu, Didi and others},
  journal={arXiv preprint arXiv:2509.23661},
  year={2025}
}

@article{wiedmann2025finevision,
  title={Finevision: Open data is all you need},
  author={Wiedmann, Luis and Zohar, Orr and Mahla, Amir and Wang, Xiaohan and Li, Rui and Frere, Thibaud and von Werra, Leandro and Gosthipaty, Aritra Roy and Marafioti, Andr{\'e}s},
  journal={arXiv preprint arXiv:2510.17269},
  year={2025}
}

@article{wang2025internvl3,
  title={Internvl3. 5: Advancing open-source multimodal models in versatility, reasoning, and efficiency},
  author={Wang, Weiyun and Gao, Zhangwei and Gu, Lixin and Pu, Hengjun and Cui, Long and Wei, Xingguang and Liu, Zhaoyang and Jing, Linglin and Ye, Shenglong and Shao, Jie and others},
  journal={arXiv preprint arXiv:2508.18265},
  year={2025}
}

@inproceedings{guo2025mammoth,
  title={Mammoth-vl: Eliciting multimodal reasoning with instruction tuning at scale},
  author={Guo, Jiawei and Zheng, Tianyu and Li, Yizhi and Bai, Yuelin and Li, Bo and Wang, Yubo and Zhu, King and Neubig, Graham and Chen, Wenhu and Yue, Xiang},
  booktitle={Proceedings of the 63rd Annual Meeting of the Association for Computational Linguistics (Volume 1: Long Papers)},
  pages={13869--13920},
  year={2025}
}

@article{zou2026advancing,
  title   = {Advancing Aesthetic Image Generation via Composition Transfer},
  author  = {Zou, Kai and Zhao, Zhiwei and Liu, Bin and Yu, Nenghai},
  journal = {International Journal of Computer Vision},
  volume  = {134},
  pages   = {252},
  year    = {2026},
  doi     = {10.1007/s11263-026-02862-8},
  url     = {https://doi.org/10.1007/s11263-026-02862-8}
}

@article{chen2025sharegpt,
  title={Sharegpt-4o-image: Aligning multimodal models with gpt-4o-level image generation},
  author={Chen, Junying and Cai, Zhenyang and Chen, Pengcheng and Chen, Shunian and Ji, Ke and Wang, Xidong and Yang, Yunjin and Wang, Benyou},
  journal={arXiv preprint arXiv:2506.18095},
  year={2025}
}

@inproceedings{onoe2024docci,
  title={Docci: Descriptions of connected and contrasting images},
  author={Onoe, Yasumasa and Rane, Sunayana and Berger, Zachary and Bitton, Yonatan and Cho, Jaemin and Garg, Roopal and Ku, Alexander and Parekh, Zarana and Pont-Tuset, Jordi and Tanzer, Garrett and others},
  booktitle={European Conference on Computer Vision},
  pages={291--309},
  year={2024},
  organization={Springer}
}

@article{chen2025blip3,
  title={Blip3-o: A family of fully open unified multimodal models-architecture, training and dataset},
  author={Chen, Jiuhai and Xu, Zhiyang and Pan, Xichen and Hu, Yushi and Qin, Can and Goldstein, Tom and Huang, Lifu and Zhou, Tianyi and Xie, Saining and Savarese, Silvio and others},
  journal={arXiv preprint arXiv:2505.09568},
  year={2025}
}

@article{arevalo2017gated,
  title={Gated multimodal units for information fusion},
  author={Arevalo, John and Solorio, Thamar and Montes-y-G{\'o}mez, Manuel and Gonz{\'a}lez, Fabio A},
  journal={arXiv preprint arXiv:1702.01992},
  year={2017}
}

@article{miao2026mibench,
  title={Mibench: Evaluating lmms on multimodal interaction},
  author={Miao, Yu and Yang, Zequn and Wei, Yake and Chen, Ziheng and Ni, Haotian and Duan, Haodong and Chen, Kai and Hu, Di},
  journal={arXiv preprint arXiv:2603.13427},
  year={2026}
}

@article{chen2024we,
  title={Are we on the right way for evaluating large vision-language models?},
  author={Chen, Lin and Li, Jinsong and Dong, Xiaoyi and Zhang, Pan and Zang, Yuhang and Chen, Zehui and Duan, Haodong and Wang, Jiaqi and Qiao, Yu and Lin, Dahua and others},
  journal={Advances in Neural Information Processing Systems},
  volume={37},
  pages={27056--27087},
  year={2024}
}

@techreport{betker2023improving,
  title={Improving Image Generation with Better Captions},
  author={Betker, James and Goh, Gabriel and Jing, Li and Brooks, Tim and Wang, Jianfeng and Li, Linjie and Ouyang, Long and Zhuang, Juntang and Lee, Joyce and Guo, Yufei and Manassra, Wesam and Dhariwal, Prafulla and Chu, Casey and Jiao, Yunxin and Ramesh, Aditya},
  institution={OpenAI},
  year={2023},
  url={https://cdn.openai.com/papers/dall-e-3.pdf}
}

@misc{flux2024,
  author={Black Forest Labs},
  title={{FLUX}},
  year={2024},
  howpublished={\url{https://github.com/black-forest-labs/flux}}
}

@article{li2024core,
  title={Core knowledge deficits in multi-modal language models},
  author={Li, Yijiang and Gao, Qingying and Zhao, Tianwei and Wang, Bingyang and Sun, Haoran and Lyu, Haiyun and Hawkins, Robert D and Vasconcelos, Nuno and Golan, Tal and Luo, Dezhi and others},
  journal={arXiv preprint arXiv:2410.10855},
  year={2024}
}

@article{stogiannidis2025mind,
  title={Mind the gap: Benchmarking spatial reasoning in vision-language models},
  author={Stogiannidis, Ilias and McDonagh, Steven and Tsaftaris, Sotirios A},
  journal={arXiv preprint arXiv:2503.19707},
  year={2025}
}

@inproceedings{aghajanyan2023scaling,
  title={Scaling laws for generative mixed-modal language models},
  author={Aghajanyan, Armen and Yu, Lili and Conneau, Alexis and Hsu, Wei-Ning and Hambardzumyan, Karen and Zhang, Susan and Roller, Stephen and Goyal, Naman and Levy, Omer and Zettlemoyer, Luke},
  booktitle={International Conference on Machine Learning},
  pages={265--279},
  year={2023},
  organization={PMLR}
}

@article{luo2024skysensegpt,
  title={Skysensegpt: A fine-grained instruction tuning dataset and model for remote sensing vision-language understanding},
  author={Luo, Junwei and Pang, Zhen and Zhang, Yongjun and Wang, Tingzhu and Wang, Linlin and Dang, Bo and Lao, Jiangwei and Wang, Jian and Chen, Jingdong and Tan, Yihua and others},
  journal={arXiv preprint arXiv:2406.10100},
  year={2024}
}

@book{gazzaniga2011cognitive,
  title={Cognitive Neuroscience: The Biology of the Mind},
  author={Gazzaniga, Michael S. and Ivry, Richard B. and Mangun, George R.},
  year={2011},
  publisher={W. W. Norton}
}

@book{stein1993merging,
  title={The merging of the senses},
  author={Stein, Barry E and Meredith, M Alex},
  year={1993},
  publisher={MIT press}
}

\clearpage

\beginappendix

\section{Data Organization Strategy}
\label{sec:data_organization}
\subsection{Data curation.}
We apply task-specific filtering to construct high-quality training data at different stages.
For multimodal pretraining, we select high-quality image--text pairs from LLaVA-OneVision-1.5-Midtraining data~\cite{an2025llava} and use Qwen3-VL-32B-Instruct to verify image--text consistency, removing samples with weak or mismatched correspondence.
For supervised fine-tuning, we further perform multiple rounds of quality control.
In particular, since the VQ-VAE tokenizer compresses images into discrete visual tokens, fine-grained visual details may be lost during tokenization, making some originally answerable samples no longer recoverable from the tokenized image.
We therefore first apply rule-based filtering and then use Qwen3-VL-32B-Instruct to verify that the visual evidence required by the target response remains available after tokenization.

\begin{table}[htbp]
    \centering
    \footnotesize
    \caption{
    Composition of the 13.7M instruction-tuning examples, including
    multimodal understanding and text-only data, together with their
    primary interplay types, including redundant information ($R$), visual-unique information ($U_v$), text-unique information ($U_t$), and synergistic information ($S$)
    }
    \vspace{2mm}
    \label{tab:training_data_composition}
    
    \setlength{\tabcolsep}{4pt}
    \renewcommand{\arraystretch}{1.20}

    \begin{tabularx}{\linewidth}{
        >{\raggedright\arraybackslash}p{2.1cm}
        >{\raggedright\arraybackslash}p{6.0cm}
        >{\raggedleft\arraybackslash}p{1.35cm}
        >{\raggedleft\arraybackslash}p{1.35cm}
        >{\centering\arraybackslash}X
    }
        \toprule
        \textbf{Primary Task}
        & \textbf{Secondary Capability}
        & \textbf{Samples}
        & \textbf{Percentage}
        & \textbf{Interplay } \\
        \midrule

        \multirow{9}{2.1cm}{
            \centering
            \textbf{Multimodal}\\
            \textbf{Understanding}\\
            \textbf{(MMU)}
        }
        & \textbf{MMU Total}
        & \textbf{9.743M}
        & \textbf{100.0\%}
        & $R$, $U_v$, $U_t$, $S$ \\

        & \quad Image Captioning \& Scene Understanding
        & 3.086M
        & 31.7\%
        & $R$, $U_v$ \\

        & \quad Knowledge-intensive Visual Understanding
        & 2.105M
        & 21.6\%
        & $U_t$, $S$ \\

        & \quad General VQA \& Instruction Following
        & 1.630M
        & 16.7\%
        & $U_v$, $S$ \\

        & \quad Object, Attribute, State \& Counting
        & 1.558M
        & 16.0\%
        & $U_v$ \\

        & \quad Scene Text \& OCR
        & 0.437M
        & 4.5\%
        & $U_v$ \\

        & \quad Visual Grounding \& Spatial Understanding
        & 0.366M
        & 3.8\%
        & $U_v$, $S$ \\

        & \quad Hallucination Suppression \& Hard Negatives
        & 0.324M
        & 3.3\%
        & $U_v$ \\

        & \quad Relational \& Compositional Reasoning
        & 0.236M
        & 2.4\%
        & $S$ \\

        \midrule

        \textbf{Text-only}
        & Text Understanding \& Knowledge
        & \textbf{3.957M}
        & \textbf{100.0\%}
        & $U_t$ \\

        \midrule

        \multicolumn{2}{l}{\textbf{Total Instruction Data}}
        & \textbf{13.700M}
        & --
        & $R$, $U_v$, $U_t$, $S$ \\

        \bottomrule
    \end{tabularx}
\end{table}

\subsection{Interplay-aware data organization.}
After data curation, we organize the 13.7M instruction examples according to four forms of task-relevant multimodal interplay:
redundant information ($R$), visual-unique information ($U_v$), text-unique information ($U_t$), and synergistic information ($S$).
Rather than exhaustively annotating every training example, we perform the analysis at the sub-task level.
For each sub-dataset, we randomly sample 100 examples and combine task-specific heuristics with model-assisted verification to characterize its dominant information structure.
Specifically, we examine whether the two modalities provide aligned or overlapping evidence, whether prediction mainly depends on the visual or textual modality, or whether both modalities are jointly required to derive information unavailable from either modality alone.
Each sampled example is scored along these criteria and assigned to its dominant interplay type.
The aggregated assignments are then used to characterize the corresponding sub-task and organize its training examples.
As summarized in \autoref{tab:training_data_composition}, this procedure provides a unified interplay-based organization across diverse multimodal understanding tasks, ranging from captioning and visual perception to knowledge-intensive reasoning and compositional inference.

The resulting instruction corpus contains 13.7M examples, including
2.8M redundant, 4.0M visual-unique, 4.2M text-unique, and 2.7M synergistic examples.
We use these categories to construct the three-stage curriculum described in Section \ref{sec:training_recipe}, progressively shifting the training emphasis from preserving pretrained capabilities, to strengthening visual perception, and finally to promoting cross-modal synergy.

\subsection{Image generation data.}
In addition to the above understanding-oriented instruction data, we collect 2.46M image-generation examples.
Since these examples primarily maintain and improve conditional image-generation capability rather than determine the interplay curriculum, we distribute them approximately uniformly across the three stages.






\section{Additional Results}
\label{sec:additional_results}

\subsection{Effect of Interplay-Oriented Curriculum Learning.}
To investigate the effect of our interplay-oriented curriculum, we conduct a controlled study starting from the same CPT checkpoint and using the same 10\% subset of SFT data. We compare two training strategies: uniformly mixing all training samples throughout SFT and progressively organizing them according to the interplay-centric curriculum. We evaluate both general multimodal capability and visual perception across different stages of training.

As shown in \autoref{tab:curriculum_result}, the interplay-oriented curriculum induces a clear shift in the capabilities learned across stages. In the early stage, where training places greater emphasis on redundant and text-unique information, curriculum learning yields substantially stronger broad visual perception, improving MME-Perception by 113.8 points over uniform mixing, while fine-grained visual perception remains weaker. As training progressively shifts toward modality-unique and synergistic information, the model becomes increasingly better at exploiting detailed and task-relevant visual cues. In particular, the curriculum initially shows a clear advantage on broad perceptual benchmarks such as MME-Perception, while its performance on MMStar gradually improves across stages and eventually surpasses uniform mixing.

These results indicate that the ordering of different interplay types affects which capabilities are emphasized during training. Earlier redundancy- and text-oriented stages favor broad perceptual capability, whereas later modality-unique and synergy-oriented stages increasingly strengthen task-sensitive visual reasoning and fine-grained perception. Overall, the curriculum does not simply provide a uniform gain across benchmarks, but progressively reshapes the model toward different aspects of multimodal capability as the information composition of the training data changes.

\begin{table*}[t]
\centering
\caption{
Comparison between uniform data mixing and interplay-oriented curriculum learning using 10\% of the SFT data.
}
\label{tab:rus_curriculum}

\fontsize{8.5}{10.2}\selectfont
\setlength{\tabcolsep}{4pt}
\renewcommand{\arraystretch}{1.05}

\begin{tabular}{lccc|ccc|ccc}
\toprule
\multirow{2}{*}{\textbf{Metric}}
& \multicolumn{3}{c|}{\textbf{Phase 1}}
& \multicolumn{3}{c|}{\textbf{Phase 2}}
& \multicolumn{3}{c}{\textbf{Phase 3}} \\
\cmidrule(lr){2-4}
\cmidrule(lr){5-7}
\cmidrule(lr){8-10}
& \textbf{Uniform} & \textbf{Curriculum} & \textbf{Diff.}
& \textbf{Uniform} & \textbf{Curriculum} & \textbf{Diff.}
& \textbf{Uniform} & \textbf{Curriculum} & \textbf{Diff.} \\
\midrule

MMStar-P
& 49.52 & 45.08 & -4.44
& 49.42 & 49.72 & +0.30
& 51.67 & 53.87 & +2.20 \\

MME-P
& 917.1 & 1030.9 & +113.8
& 1135.8 & 1178.5 & +42.7
& 1249.5 & 1211.6 & -37.9 \\

MMVP
& 26.67 & 28.00 & +1.33
& 32.67 & 26.00 & -6.67
& 29.33 & 31.33 & +2.00 \\

CVBench-2D
& 68.64 & 68.08 & -0.56
& 73.02 & 72.11 & -0.90
& 73.99 & 74.62 & +0.63 \\

\bottomrule
\end{tabular}
\label{tab:curriculum_result}
\end{table*}

\begin{table}[t]
\centering
\caption{
Evaluation on remote-sensing benchmarks.
\textbf{RS Data} denotes the number of remote-sensing samples used for fine-tuning.
}
\label{tab:rs_result}

\setlength{\tabcolsep}{6pt}
\renewcommand{\arraystretch}{1.05}

\begin{tabular}{lccc}
\toprule
\textbf{Model}
& \textbf{RS Data}
& \textbf{FGRS}
& \textbf{FGRC} \\
\midrule

BAGEL
& -- 
& 42.51
& 36.65 \\

\midrule
GeoChat
& -- 
& 53.47
& 21.90 \\

SkySenseGPT
& 1.415M
& \textbf{79.76}
& \textbf{55.50} \\

\midrule
Lumina-DiMOO$^\ast$
& 50K
& 77.77
& -- \\

Gestalt
& 50K
& \underline{79.34}
& \underline{38.20} \\

\bottomrule
\end{tabular}

\vspace{2pt}
{\footnotesize $^\ast$ FGRC evaluation did not complete successfully.}
\end{table}

\begin{figure}[t]
\centering
\includegraphics[width=0.85\linewidth]{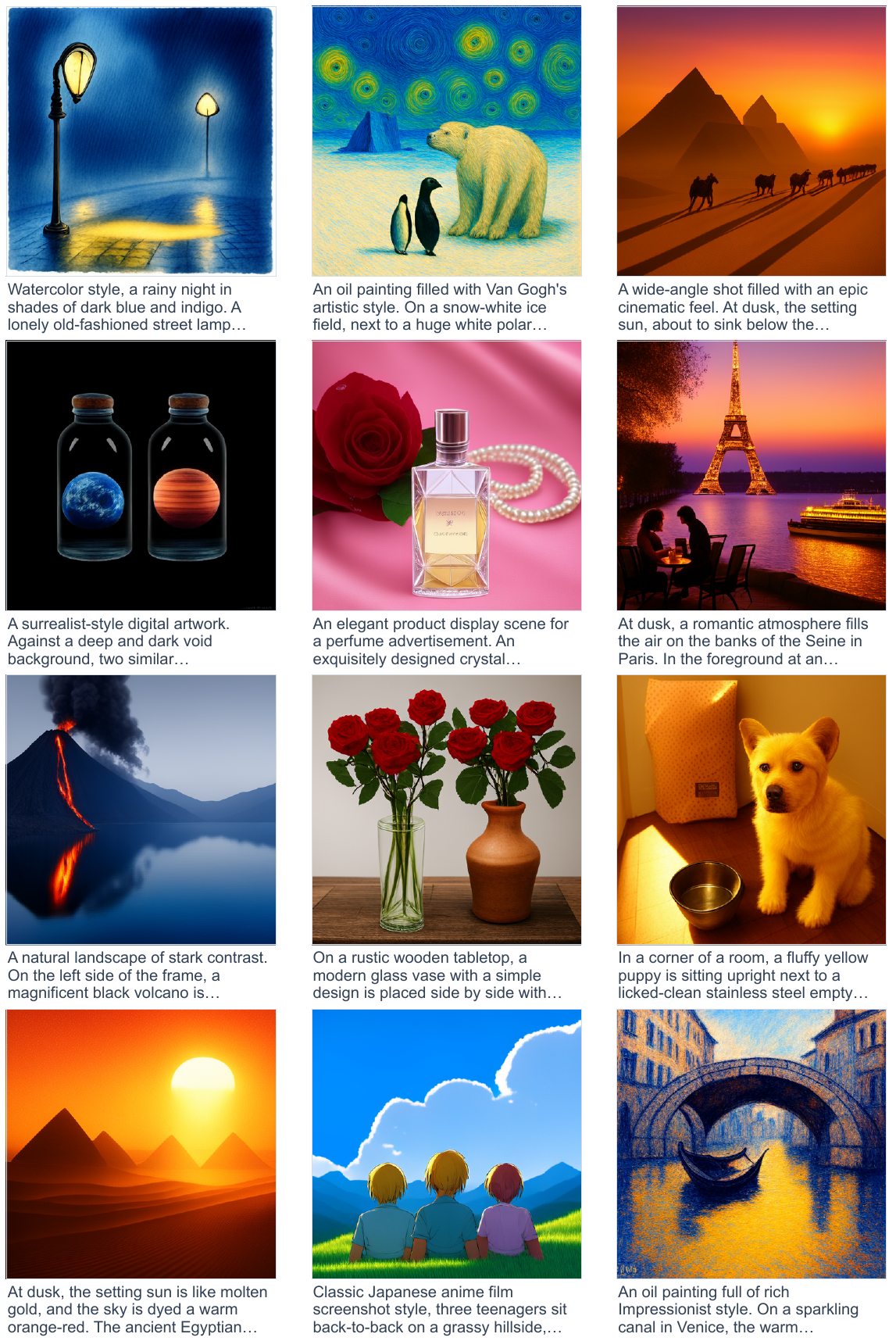}
\vspace{-0.5em}
\caption{Qualitative results on UniGenBench. Gestalt demonstrates strong fine-grained generation capabilities on complex prompts involving multiple objects, attributes, and spatial relations.}
\vspace{-1em}
\label{fig:qualitative_unigen}

\end{figure}

\subsection{Adaptation to Remote-Sensing Tasks.}
To examine whether Gestalt can efficiently adapt to a substantially different visual domain, we introduce only 50K remote-sensing VQA samples from FIT~\cite{luo2024skysensegpt} into the training mixture and evaluate the resulting model on FGRS and FGRC. The two benchmarks emphasize complementary aspects of remote-sensing understanding: FGRS evaluates fine-grained recognition and understanding of remote-sensing imagery, whereas FGRC places greater emphasis on reasoning about relations among objects and regions. To provide a controlled comparison in terms of domain-specific data exposure, we additionally adapt the official Lumina-DiMOO checkpoint using the same 50K remote-sensing samples, denoted as Lumina-DiMOO$^\ast$.

As shown in \autoref{tab:rs_result}, Gestalt achieves 79.34 on FGRS, substantially outperforming the general-purpose BAGEL baseline and GeoChat, and approaching the 79.76 performance of SkySenseGPT despite being exposed to only 50K remote-sensing samples, compared with 1.415M for SkySenseGPT. Under the same 50K domain-specific data exposure, Gestalt also surpasses Lumina-DiMOO$^\ast$ on FGRS. On the more relation-oriented FGRC benchmark, Gestalt further improves over BAGEL and GeoChat, although a gap remains compared with the heavily specialized SkySenseGPT. These results show that Gestalt can effectively absorb domain-specific visual knowledge from limited additional data while retaining strong general multimodal capabilities, demonstrating its adaptability and data efficiency when extending to new domains.



\subsection{Qualitative Analysis of Fine-Grained Generation.}
\autoref{fig:qualitative_unigen} presents qualitative examples on UniGenBench, covering diverse visual styles, multi-object compositions, attribute constraints, and spatial relations.
Across these elaborate and challenging prompts, Gestalt demonstrates strong instruction-following ability while maintaining coherent scene composition and visual quality.

In particular, Gestalt can faithfully organize multiple semantic constraints within a single image.
For example, the generated scenes preserve clear object identities and relative layouts in complex compositions, such as the polar bear and penguins on a snowy field, the camel caravan against the pyramids, the couple and cruise ship along the Seine, and multiple flower arrangements with distinct containers.
These examples require more than generating the requested objects independently: the model must correctly bind attributes to their corresponding entities and arrange them according to the specified scene structure.
Gestalt also handles fine-grained object properties, as illustrated by examples involving similar bottles with different internal contents, flowers placed in different types of vases, and a puppy positioned beside an empty metallic bowl.

Meanwhile, Gestalt exhibits strong controllability over visual style and global appearance.
The generated results span substantially different styles, including watercolor, Van Gogh-inspired oil painting, Impressionism, surreal digital art, product photography, cinematic landscapes, and Japanese anime, while preserving the semantic content of the corresponding prompts.
Notably, stylistic constraints affect the entire visual composition rather than appearing as superficial local textures: lighting, color palette, brushwork, perspective, and scene atmosphere remain mutually consistent within each image.
The watercolor night scene, warm cinematic desert landscapes, and Impressionist canal scene provide particularly clear examples of such global stylistic coherence.

Overall, these qualitative results highlight Gestalt's ability to jointly model
\emph{what} should appear, \emph{where} different elements should be placed, and
\emph{how} the resulting scene should be rendered.
This is especially important for fine-grained text-to-image generation, where success depends not only on visual realism but also on preserving compositional and attribute-level constraints from the input instruction.
Together with the quantitative results on UniGenBench, the examples suggest that Gestalt can effectively translate detailed textual specifications into coherent visual structures across a wide range of generation scenarios.

\end{document}